\documentclass{article}

\usepackage[preprint, nonatbib]{neurips_2023}

\usepackage[utf8]{inputenc} 
\usepackage[T1]{fontenc}    
\usepackage{xr-hyper}
\usepackage[hidelinks]{hyperref}       
\usepackage{url}            
\usepackage{booktabs}       
\usepackage{amsfonts}       
\usepackage{nicefrac}       
\usepackage{microtype}      
\usepackage{graphicx}
\usepackage{fancyvrb}
\usepackage{xcolor}         

\title{KG-FRUS: a Novel Graph-based Dataset of 127 Years of US Diplomatic Relations}

\author{%
  {Gökberk Özsoy$^1$, Luis Salamanca$^1$, Matthew Connelly$^2$,}\\
  {\textbf{Raymond Hicks$^2$ and Fernando Pérez-Cruz$^1$}}\\
  1. Swiss Data Science Center,
  ETH Zürich,
  Zürich, 8092, Switzerland\\
  \texttt{gokozsoy20@gmail.com, \{luis.salamanca, fernando.perezcruz\}@sdsc.ethz.ch} \\
  2. History Lab,
  Columbia University,
  New York City, 1027, USA\\
  \texttt{\{mjc96, rh2883\}@columbia.edu} \\  
}

\begin{document}

\maketitle

\begin{abstract} 
In the current paper, we present the KG-FRUS dataset, comprised of more than  300,000 US government diplomatic documents encoded in a Knowledge Graph (KG). We leverage the data of the Foreign Relations of the United States (FRUS) (available as XML files) to extract information about the documents and the individuals and countries mentioned within them. We use the extracted entities, and associated metadata, to create a graph-based dataset. Further, we supplement the created KG with additional entities and relations from Wikidata. The relations in the KG capture the synergies and dynamics required to study and understand the complex fields of diplomacy, foreign relations, and politics. This goes well beyond a simple collection of documents which neglects the relations between entities in the text. We showcase a range of possibilities of the current dataset by illustrating different approaches to probe the KG. In the paper, we exemplify how to use a query language to answer simple research questions and how to use graph algorithms such as Node2Vec and PageRank, that benefit from the complete graph structure. More importantly, the chosen structure provides total flexibility for continuously expanding and enriching the graph. Our solution is general, so the proposed pipeline for building the KG can encode other original corpora of time-dependent and complex phenomena. Overall, we present a mechanism to create KG databases providing a more versatile representation of time-dependent related text data and a particular application to the all-important FRUS database.
\end{abstract}

\section{Introduction} 

The vast text availability nowadays has enabled wondrous technical achievements, and the capabilities of modern large language models are their best example. Yet, this is still a snapshot of all produced text throughout history. Efforts in digitalizing historical archives have generated valuable corpora, especially for social sciences. In the best-case scenario, these documents are encoded as queryable free text files and are conveniently curated and annotated. However, important information is not encoded in the documents but rather in the relations between them and the entities within. By capturing dynamics and synergies within the entities, the connections spanned by these relations can help to study more complex phenomena, discover hidden patterns, and gain a more profound understanding of the constituent elements.
 
We present KG-FRUS, a graph-based dataset that describes the cables over 120 years of US diplomatic history. To build it, we have leveraged the work published in \cite{frusrepo}, in which the declassified documents from the Foreign Relations of the US (FRUS) have been encoded into XML files. While this representation enables the exploration of these documents, the complex fields of geopolitics and diplomacy rely upon understanding the interaction between international actors, and how these evolve through time. These limitations drove the creation of KG-FRUS, which focused on: building a flexible dataset around an expandable knowledge graph (KG), carefully annotated with time stamps to capture time dynamics, and linked to the original transcripts for future enrichment. For its creation, we first need to identify and understand the entities to extract (from the text itself and associated metadata) and build a base KG schema from them. We can expand the KG by using the nodes, relations, and their attributes, to link it to existing structured databases, such as Wikidata \cite{wikidata}. The KG schema can be further enriched by using different natural language processing (NLP) methods on the linked documents’ transcripts, to parse and learn extra information from them. 
 
To showcase the possibilities of KG-FRUS, we probe the data using different methods: from simple queries to illustrate varied statistics and performing knowledge discovery to more complex graph algorithms to fully exploit the vast dimension of the KG. This way, we intend to emphasize that this data representation is not only a more exhaustive and complete representation of the original corpora and the information contained within. It is also a more accessible and useful representation for a broader audience. For example, experts can manually explore the data by asking domain-specific research questions \cite{claimskg}, and data-driven researchers could leverage the KG through different graph algorithms \cite{abu2021relational}. Besides, machine learners can resort to the availability of large text, and large and complex graph structures, to develop and benchmark methods in NLP, graph machine learning, and related fields.
 
The contribution of this paper is twofold. First, we provide a knowledge graph representation of the Foreign Relations of the US documents. Second, we comprehensively provide the steps needed to create the KG from any other national and international archives.

\section{Related Work}\label{related_work} 

In social and political sciences, researchers commonly use off-the-shelf NLP methods to analyze the text in hand to capture different patterns. However, KG use is seldom used, but pioneering works have proved its potential. In this section, we will explore these two directions in more detail. US Congressional Record spans from 1858 to 1994 and includes six million speeches by US Members of Congress. It is a huge dataset, and authors use corpus-specific word embeddings to analyze emotionality change over time depending on political party, gender, and era, among others \cite{gennaro2022emotion}. State Department records from 1973-1979 transmitted to and from Iran are used to conclude that officials in Iran had reported on the protests leading to Islamic Revolution but US officials could not comprehend the severity of the situation \cite{iranpaper}. The authors of this paper rely on basic traffic statistics and a dictionary-based sentiment analyzer during their analysis. The United Kingdom Government Web Archive \cite{ukarchives} holds the central government information from 1996. Researchers have developed a method for efficient semantic search in this archive \cite{ibmpaper}. They first extract named entities within each text, then train document embeddings via doc2vec \cite{doc2vecpaper}, which enables searching on entities or fetching similar documents for a topic. Similarly, in \cite{dynamicpartypaper}, document embeddings are used to capture the dynamic shift of political actors through time. In their formulation, the party-related documents for a given time are considered as one. The document embedding can be treated as political party embedding. In this way, they could study ideology shifts reflected by party embeddings. They also compared them to keywords such as abortion or immigration. They use congressional records for the United States, United Kingdom, and Canada.

An early effort for political KGs is the BBC Politics Ontology \cite{bbcpoliticsontology}. This ontology is designed for the United Kingdom local government elections and European Parliament Elections in May 2014. POLARE \cite{polarepaper} is concerned with producing a KG about political agents in Brazil. The main data source is the House of Deputies and Senate but also relies on crowd-sourced contributions. This KG relies on significant manual labor, and it is not scalable. It focuses on people, organizations, and their connections in Brazil but also captures legislative aspects and electoral processes. In another work, authors compile a fusion KG obtained from heterogeneous sources including official sources about Australian Members of Parliament, and various Australian Twitter users that comment on politics, WordNet, and Politics domain ontology \cite{abusalihpaper2}. This KG produces embeddings showing how people with similar political ideologies are closer (Euclidean distance) in the embedded space.

\section{KG-FRUS}\label{construction}

To build KG-FRUS, we leverage the FRUS corpus (publicly available in digital format \cite{frusrepo}). Each of its XML files represents a single volume, and they have been prepared according to Text Encoding Initiative (TEI) P5 guidelines, as well as following some project-specific encoding guidelines and conformance requirements\footnote{https://github.com/HistoryAtState/frus/blob/master/schema/frus.odd}.

Therefore, FRUS is a collection of volumes, each being the concatenation of documents, i.e., declassified historical records, on a particular topic for a certain president’s term. Despite the importance of relations between documents and entities within them, this data representation does not encode them and hence does not allow harnessing their possibilities. We see this as the most significant weakness in FRUS. This limitation also applies to most other datasets in social and political sciences. It is the main pitfall for the adequate study and understanding of social phenomena, historical events, etc. For these reasons, we advocate that a knowledge graph provides a richer, more flexible, and exhaustive representation of the data. The KG can be better exploited in downstream tasks and by a larger set of researchers.

In the present section, we illustrate the process carried out to create KG-FRUS from FRUS. In a nutshell, this requires creating a schema for the KG from the documents’ entities and metadata and using it to populate the graph. This schema can be expanded first by relating or adding nodes or relations using existing ontologies and/or structured databases. Then, by analyzing the text transcripts employing different NLP methods, we can further enrich the graph with new nodes and attributes. This procedure can obviously be applied to many corpus/datasets in various fields, allowing us to identify a general paradigm.

\subsection{Parsing and Base Graph Population}\label{parsing_population} 




Each data source requires an ad-hoc understanding of the entities contained and related metadata. We need to explore it to build the schema and parse it appropriately. FRUS is built around XML files, and we illustrate the common hierarchy of files in Fig.~\ref{fig:xmlhierarchy}. Only small variations occur of this structure across files. 


\begin{figure}[ht]
\centering
\begin{small}
\begin{Verbatim}[commandchars=\\\{\}]
<?xml version="1.0" encoding="UTF-8"?>
<TEI xml:id="frus1969-76v30" 
    <teiHeader>\textcolor{blue}{<fileDesc>}
    </teiHeader>
    <text>
        <front>\textcolor{blue}{<terms><persons>}
        </front>
        \textcolor{blue}{<body>}
            <div type="compilation" xml:id="comp1">
                \textcolor{blue}{<div type="document" xml:id="d1">}
                    \textcolor{blue}{<head><persName><placeName><signed>}
                </div>
                other documents...
            </div>
        </body>
        <back>...
        </back>
    </text>
</TEI>
\end{Verbatim}
\end{small}
\caption{XML hierarchy of FRUS volumes, with the main tags employed highlighted in blue.}
\label{fig:xmlhierarchy}
\end{figure}

The main identifier of the volume and various publication information is contained in children tags of \verb|<teiHeader>|. Next, among the children tags contained in \verb|<front>|, we parse \verb|<terms>| and \verb|<persons>|, as they hold annotation id, name, and description information of all terms and people mentioned in the volume. The tag \verb|<body>| of type ``document'' holds the transcripts' texts, which can be of subtype ``historical-document'' or ``editorial-note'' (notes explaining particular events in the course of a particular era). Children tags of \verb|<body>| contain most of the information required to populate the KG. Among the most relevant ones, \verb|<head>| holds the title of document, chapter, or compilation, \verb|<persName>| and \verb|<gloss>| hold person and term ids, respectively, whose annotation information is given in \verb|<front>|. \verb|<placeName>| is the city the document was sent from, and \verb|<signed>| holds the signed person's name. For further details, in Appendix A we provide a thorough description of all remaining tags, and how they have, or have not, been utilized. 

\begin{figure}[ht]
\begin{center}
\includegraphics[width=\linewidth]{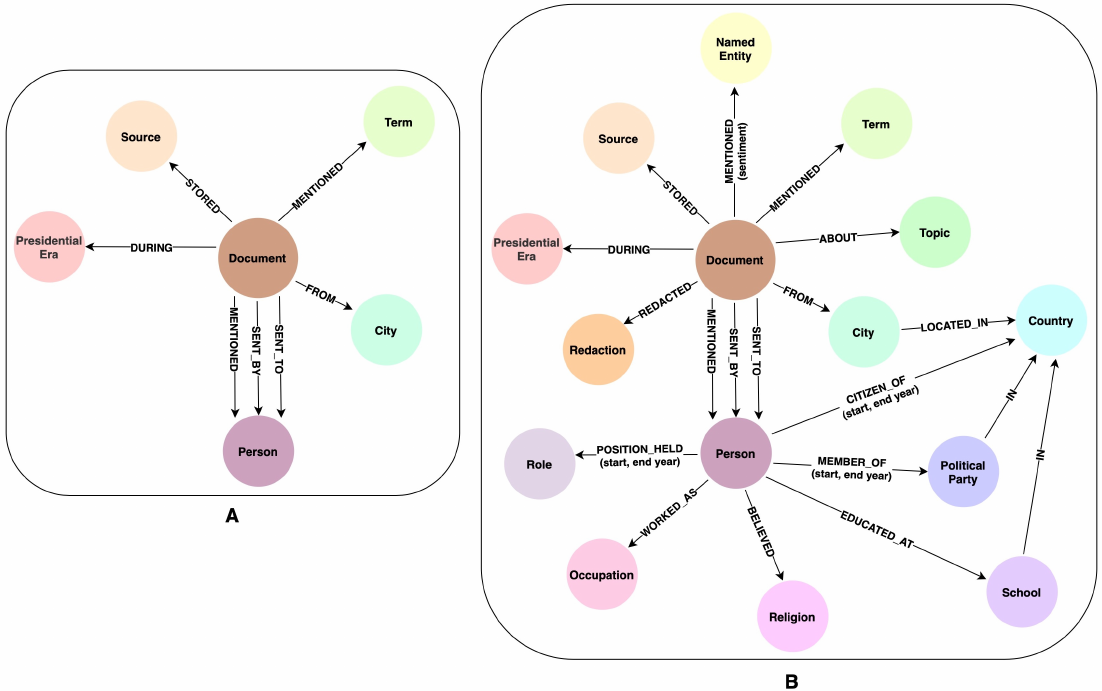}
\end{center}
\caption{Base and Expanded Knowledge Graph for KG-FRUS.}
\label{fig:schema}
\end{figure}

The knowledge drawn from the previous exploration allows us to propose the schema illustrated in Fig.~\ref{fig:schema}A for a ``Base KG''. The core node is ``Document'', described by the attributes \textit{docId}, \textit{subtype}, \textit{date}, \textit{year} and \textit{volume}. It is linked to a ``Presidential era'' (obtained from the date), ``Source'' (where the document is stored physically), ``City'', to all ``Term'' nodes annotated, and all ``Person'' instances contained, either in the main text (through the relation MENTIONED), or refer to in the \verb|<body>| header as the sender or receiver (relations SENT\_BY and SENT\_TO respectively). Further information on the institution of the sender and receiver, e.g.~Embassy in Ankara or White House, is stored as an attribute of the corresponding ``Document''. Further details on the tags employed to build each of the aforementioned nodes, relations, and attributes are described in Appendix A.

Once defined the KG schema and the tags related to each node, and their attributes, we finally iterate over all XML files to extract the information required\footnote{For parsing, we use Python's standard library \href{https://docs.python.org/3/library/xml.etree.elementtree.html}{ElementTree XML API}} and populate the ``Base KG''. The technology chosen for the KG is Neo4J \cite{neo4j}, and we rely on native tools for populating the graph through the CSV files containing all the information. For the case of the documents' text, we could store it as an attribute of the ``Document'' node. However, that would hinder the performance when querying the graph. For that reason, we opted for an accessory SQL database, where a single table stores the text, together with the \textit{docId} that links it back to the KG.  

\subsection{Expanded KG construction}  


One of the best advantages of the proposed approach is the complete flexibility offered by a KG. It enables to easily add and remove nodes, create new relations, and freely use attributes. We now harness this possibility to perform the following steps. First, we expand the graph using the nodes' attributes. Later, we resort to NLP methods and the transcript, i.e.~free text, related to each ``Document'' node to further enrich the KG. We don't intend to perform exhaustive exploitation of the data but rather illustrate the possibilities of the paradigm presented, and to compute additional nodes/relations that will be exploited in further downstream tasks. The final ``Expanded KG'' obtained is depicted in Fig.~\ref{fig:schema}B.

\subsubsection{Expansion through entities}\label{entityexpansion}

In order to correctly rely on the created nodes, and associated attributes, for the expansion of the graph, we need to bring coherence to them. This is the case for the ``Person'' and ``Term'' nodes, as in both cases the same instance can be referred to through different formulas. We mostly tackle the problem of ``Person'' unification, of major importance for the KG envisaged. Still, in Appendix A, we provide details on the efforts carried out to also unify the ``Term'' node.

\paragraph{Person Unification}Here we tackle the problem of grouping all names, across volumes and documents, that refer to the same person. For example, President Richard Nixon can be found as "Richard M. Nixon", "Richard Milhous Nixon", "Richard Nixon", and "Nixon Richard". This is without considering possible typos that might exist in the text. Besides, as FRUS editors confirm, identifiers are useful only for single-volume purposes, and no coherence across volumes exists. To solve this issue, we implement the following unification algorithm where, after each merging step, all matched names, ids, and descriptions are concatenated. The merging criterion at each step is:

\begin{enumerate}
    \item Exactly matched names (to unify ids across volumes).
    \item Names with the same words but ordered differently, e.g.~"Richard Nixon" and "Nixon Richard". 
    \item At least two common words, plus either a Damerau-Levensthein distance of at most 5 or Jaro similarity of at least 0.9 (to reduce near-duplicate names).
    \item A Damerau-Levensthein distance of at most 1 (for the case of obvious misspellings).
\end{enumerate}

While steps 1 and 2 cannot lead to any wrong merging, 3 and 4 can generate false positives. We reduce the likeliness of this by adjusting the distances, hyperparameters of the method, to seemingly conservative values. 
For each unique person obtained, we have associated lists of unified names, ids, and descriptions. This information is used to create a single ``Person'' node (instead of many), which holds all the lists with the provenance, and preserves all relations of the original names\footnote{For the sake of simplicity, the merging is performed before the population in Section \ref{parsing_population}}. 

{In the literature, this problem is normally known as entity resolution, and can be applied more broadly to other categories, such as numbers, emails, acronyms, etc. We have opted to refer to it as person unification. First, because we just focus on names. And secondly, because this problem was initially motivated by the need of merging person's names across volumes that, even though match exactly, had different ids in the volumes' metadata.}
  
  


\paragraph{Person Wikification}\label{par:entitylinking}We link each person with the corresponding Wikidata \cite{wikidata} entry, if exist. For each unique ``Person'' node, we employ the associated name list to query Wikidata using SPARQL \cite{sparql}. However, for some common names, multiple matches occur. In this case, we use Sentence BERT \cite{sbert} to disambiguate. We compare the FRUS embedding of the person, obtained by averaging the embeddings of its list of descriptions, against the embedding computed from the header text of each Wikidata candidate. Then, the Wikidata entry with the highest cosine similarity is selected. Beyond, different ``Person'' nodes associated with the same Wikidata entry are also merged together. {It is important to bear in mind that the wikification process is not flawless, and may induce some errors, mostly in the case of more uncommon names, as explained in Appendix C.}

Associating ``Person'' nodes to Wikidata enables us to use the structured data contained in it. As illustrated in Fig.~\ref{fig:schema}B, the extracted fields are added either as new nodes, relations, or attributes. These are gender, religion, school, occupation, positions held (i.e. role), citizenship, country, and political party membership. Many of these include specific starting and ending dates, which are encoded in the corresponding relations, allowing to add a dynamic character to the KG.  Finally, given the new node ``Country'', we perform a final step of city-country matching. We devise a hybrid, automatic and manual, algorithm that relies on a city-country database \cite{worldcities} to automatize the matching, requesting only manual intervention when several matches are found (e.g. Sucre in Colombia and Bolivia). More details are provided in Appendix A. 


\subsubsection{NLP enrichment}\label{nlpenrichment}

\paragraph{Named Entity Recognition}\label{Named Entity Recognition}In the FRUS corpus, only person's names and institutions are annotated. However, persons (beyond the ones already annotated), nationalities, locations, organizations, or events mentioned within the documents can provide further insights. We extract these entities with Spacy NER \cite{spacy2}, excluding date, time, quantity, ordinal, cardinal, money, and percent values. A new node ``Named Entity'' is created to contain these, with an attribute indicating the specific type. Currently, person's names and terms are not unified with those already annotated, and we left this task for future releases of the dataset.

\paragraph{Redaction Extraction}\label{par:redactionextraction}Omitted text that remains classified after declassification is called redaction. Redactions are important as they tend to hide critical intelligence actions such as a person, decision, money amount, or place name. In FRUS, redactions are in the format of italic string in between bracketed insertions, e.g. \verb|[<hi type="italic">1 line not declassified.<\hi>]|. They can be of different types (monetary, name, place, excerpt of text, etc.), and reflect details on the amount of material hidden. Given that this text is semistructured, these types, as well as any further details, are extracted through a combination of POS tagging and sequence matching, as detailed in Appendix A.

\paragraph{Topic Modeling}\label{par:topicmodeling}This technique provides a general view of the semantic landscape, shows the most frequent topics and interconnects documents across different years. We use both LDA \cite{lda} and BERTopic \cite{bertopic} in the extracted documents. With BERTopic we compute two sets of topics. First, for a maximum of 300 and using the original documents, leads to topics mostly related to specific geographical locations, historical events, and/or people. Then, to obtain general topics agnostic to specific events, we feed the documents with the NER named entities removed, and reduce to a maximum of 100. LDA is restricted to a maximum of 50 topics. The results of the 3 models are added to the graph as 3 new nodes, which can be utilized depending on the specific domain questions. Thanks to the flexibility offered by the KG, new case-specific nodes could be added if required.


\subsection{Database creation and workflow}

To advocate for reproducibility in research, especially in the process of data manipulation and creation, we have leveraged Renku \cite{renku} to track the dataset creation\footnote{Link to the repository \url{https://renkulab.io/projects/luis.salamanca/kg-frus}}. This technology allows, beyond standard code versioning, tracking data and workflows. This enables to visually understand the data flow throughout the different pipeline steps, from the input data to the intermediate results, final KG-FRUS files and plots generated from it. More details are provided in Appendix B and \cite{renku2}.

\subsection{KG-FRUS Statistics and Validation}


KG-FRUS is the result of analyzing the 542 volumes of the original FRUS dataset \cite{frusrepo}, containing a total of 311,604 diplomatic documents from 1861 to 1988. The KG populated using this data contains 812,477 nodes, being the majority the 311,604 ``Document'' nodes. These are grouped into 26 ``Presidential Era'', related to  18,409 annotated and unified ``Person'' nodes. Among many other nodes and connections, ``Person'' nodes are linked to for example 801 ``Occupation'' and 3256 ``Role'' nodes, as extracted from Wikidata. Besides, more than 9,000,000 relations, of 16 types, connect the nodes.


To avoid misinterpreting results, it is crucial to keep in mind the following statistics, as they relate to the completeness of the KG and the extraction performance.
First, we need to highlight that only in 271 volumes, out of the total 542, person's names are annotated, mostly from the Eisenhower era (1953). Using these annotated entities for the process of person unification described in Section \ref{entityexpansion}, we start with a total of 60,740 names, which are reduced through Steps 1 and 2 to 24,187 ($40\%$ of the initial count), which shows that the annotations are actually tidy and near-unified. Then, through Steps 3 and 4, the count is reduced further to 18,713, an extra $10\%$. For these steps, we have chosen conservative values for the distances, in order to avoid false positives. To validate this process, we have randomly selected 423 names unified in Steps 3 and 4, and assess if the merging is carried out correctly, or if there is any mismatch, i.e. it is not assigned to the correct list of person's names. Only 37 were wrong, leading to an accuracy of $91.2\%$. Finally, by merging ``Person'' nodes with the same Wikidata entry, we reduce the count further to 18,409. In conclusion, this process enables reducing all annotations, not unique in the original FRUS, to only a $30\%$ of the original count, essential to carry out precise studies leveraging on the international actors mentioned throughout the corpus.

\begin{figure}[ht]
\centering
     \includegraphics[width=0.6\textwidth]{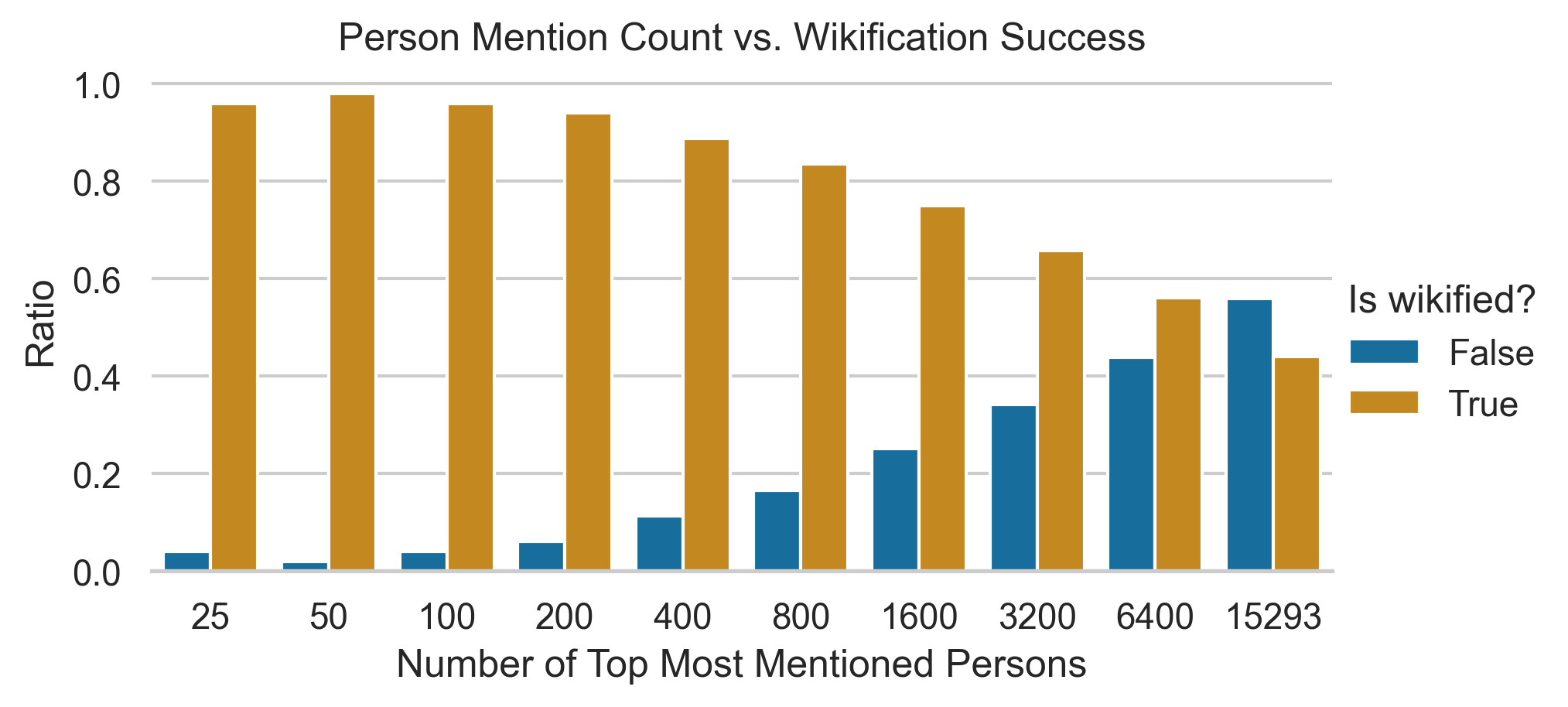}
      \caption{Ratio of ``Person'' nodes associated to its Wikidata entry, ordered by the number of times they are mentioned in the documents.}
       \label{fig:wikification success}
\end{figure}

For person \textit{wikification}, we expect frequently mentioned people to be the most historically prominent ones as well, and therefore the ones easiest to link to their corresponding Wikidata entry. We illustrate this in Figure \ref{fig:wikification success}, where the top most mentioned person's names are linked with higher success, gradually diminishing as we include more infrequent ones. Out of the 466,253 MENTIONED relations, $80\%$ correspond to ``Person'' nodes associated with a Wikidata entry, and therefore expanded with the additional nodes listed in Section \ref{entityexpansion}, underlining the validity of this method for further historical analysis. To validate the \textit{wikification} process, we selected 100 random ``Person'' nodes, and assess the correctness of the Wikidata entry associated. $90\%$ are correct, having this group also a larger mean MENTIONED count ($48.38$ vs $28$), which signifies how the automatic assignment mostly fails with person's names appearing less in the text.

The previous numbers and figures are just an overview of the magnitude of KG-FRUS. The complete statistics are provided in Appendix C, with different tables and figures containing exhaustive information.

\section{Applications}\label{applications}

The chosen data structure allows a wide range of approaches and methods to probe and utilize the data depending on the application and/or research question. A historian can explore the graph, by manually expanding the relations, or by using some specific queries. A data scientist can apply graph algorithms to, for example, learn node embeddings, and perform clustering from them. Or the full-scale of the information encoded can be leveraged to perform KG augmentation, in order to predict missing or hidden relations and patterns. In the current section, we briefly showcase some applications, just as a small preview of how KG-FRUS could be used in different domains, applications, and downstream tasks. In Appendix D we provide expanded results to each of these sections. 

{It is important to bear in mind that the applications presented are just a glimpse of what could be done leveraging KG-FRUS, and we know larger and more involved analyses could be carried out. But showing these analyses is not the focus of the paper, but rather encouraging domain scientists to pose interesting research questions that allow fully exploiting KG-FRUS possibilities. Similarly, we need to stress out that the results presented are not motivated by any specific research question, and are mostly showing well-know historical knowledge, in hindsight helping to validate KG-FRUS and the presented analyses.}


\subsection{Timeline of missives' origins}

\begin{figure}[htb]
    \centering
     \includegraphics[width=1\textwidth]{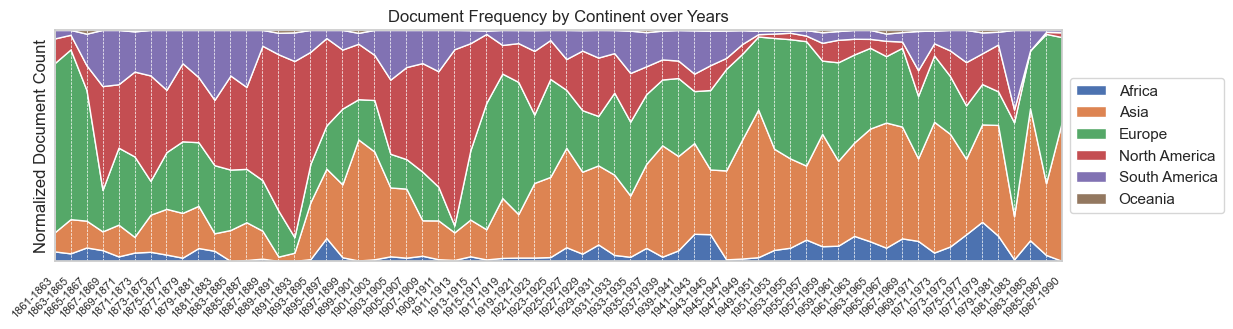}
     \vspace*{-0.5cm}
      \caption{Stacked document count by source continent over years (excluding documents originated from US), in periods of 2 years.}
       \label{fig:doc count by continent}
\end{figure}

By querying the KG to count documents' continent of origin, we can illustrate how US diplomacy has steered its attention throughout time depending on specific events and locations. In Figure \ref{fig:doc count by continent} (the values are normalized by the total number of documents for periods of 2 years), we can discern specific events, such as both World Wars, the whole tension during the Cold War, the Falkland war in 1982 (peaks for Europe and South America), and see how, naturally, Europe and Asia particularly draw greater attention. We can go as fine-grain as we want, and we believe this type of exploration could be utilized to unveil how an increase in the frequency can pinpoint the emergence of specific events and/or phenomena.

\subsection{Redaction Analysis} 


In FRUS, redaction types are diverse, which allows varied investigations on different aspects of the documents and the information hidden. To illustrate how we can leverage the extracted redactions, in Table \ref{table:bertopic dollaramount count} we show the 5 topics with the largest number of redactions of monetary type. Intelligence officials might not want to reveal the amount of money invested in certain policies, hence leaving these fields classified. Entities mentioned in topic descriptions contained names and locations that relate to possible events, which may explain the reasons for keeping this information hidden.



\begin{table}[ht]
\begin{center}
\caption{Most redacted 5 topics in terms of dollar amount redaction. Topic description by most prominent 7 words. Topic model is BERTopic with named entities.}
\resizebox{0.8\textwidth}{!}{%
\begin{tabular}{ p{11cm} p{1.4cm} p{1.4cm}  }
\toprule
\vspace*{0.01cm}Topic Description & {Document Count} & Redaction Count \\
\midrule
chile - latin - allende - somoza - costa - nicaraguan - american  & 56 & 453\\
united states - secretary - general - soviet - chinese - ambassador - position & 46 & 200\\
tshombe - mobutu - south african - lumumba - support - belgians - rhodesian &48& 135\\
scowcroft - portugal - potts - kissinger - mr colby - dollar - president spinola 	&28	&125\\
nsc - central intelligence - usia - cia - agencies - security - secretary dulles 	&6	&18\\
\midrule
\end{tabular}
}
\label{table:bertopic dollaramount count}
\end{center}
\end{table}

\subsection{Dynamic Entity Embeddings}\label{Dynamic Entity Embeddings} 
\begin{figure}[ht]
\centering
     \includegraphics[width=0.8\textwidth]{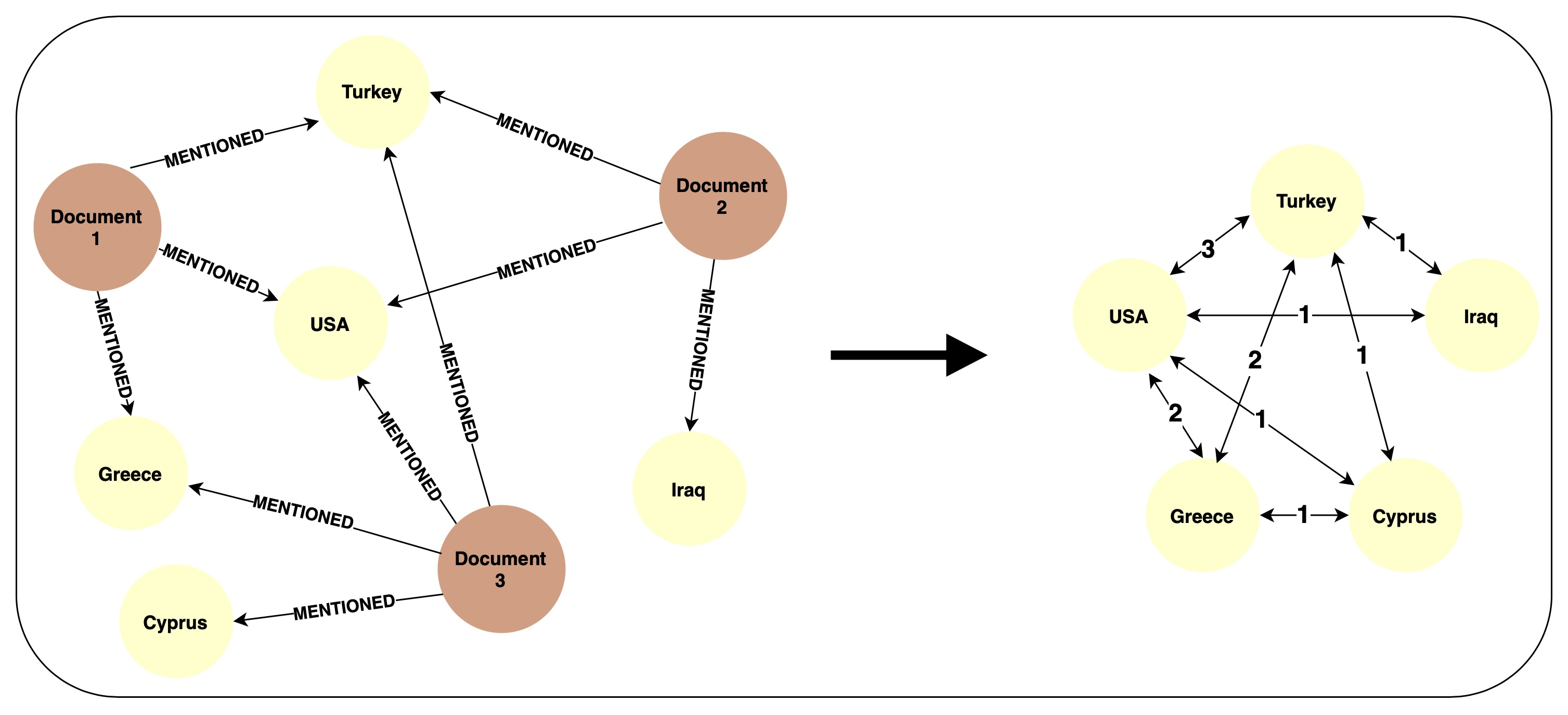}
      \caption{Conversion of ``Document''-``Named Entity'' subgraph into co-occurrence graph.}
       \label{fig:dynamic entity embeddings}
\end{figure}

Countries steer their foreign policies over time due to war, collaborations, economic interests, etc. To probe if these phenomena are captured by how relations between entities change, we propose to use dynamic entity embeddings, i.e. captured for specific time windows. This allows comparing how they get closer or further away depending on the specific world events, and how this reflects specific entity's political shifts. To obtain the embeddings, we first restrict the graph to the range of 1949 to 1985 and remove the ``Named Entity'' nodes that occur less than 50 times. The subgraph used is formed by the ``Document'' and ``Named Entity'' nodes, related through the MENTIONED relation. As illustrated in Fig.~\ref{fig:dynamic entity embeddings}, this is reduced to a unipartite graph, where the newly created relation encodes the number of times the related nodes have co-occurred. Finally, we utilize the implementation of Node2Vec \cite{node2vec} and FastRP \cite{fastrp} algorithms included in the Neo4J suite, which allows efficient computation of these vectors. 

In Table \ref{table:portugal}, we present the most similar entities over time to Portugal. Busy with Europe initially, it shifts its attention to African colonies during the late 1950s and 60s, probably due to the decolonization era. Afterward, it mostly deals with global issues such as communism, and OPEC. Similar results are presented for NATO and Gibraltar in Appendix D.

\begin{table}[htb]
\caption{Ten most similar entities to Portugal over years.}
\centering
\resizebox{\textwidth}{!}{%
\begin{tabular}{p{2.1cm}p{2.1cm}p{2.1cm}p{2.1cm}p{2.1cm}p{2.1cm}p{2.1cm}p{2.1cm}p{2.1cm}}
\toprule
49 to 53 & 53 to 57 & 57 to 61 & 61 to 65 & 65 to 69 & 69 to 73 & 73 to 77 & 77 to 81 & 81 to 85\\
\midrule
Norwegian &Scandinavian &Sweden &Portuguese &Lisbon &Portuguese &Democrats &US &Turkey \\
Belgium &Belgium &Norway &South Africans &Africa &United Kingdom &Atlantic &Europeans &Cyprus \\
Netherlands &Denmark &Commonwealth &Azores &Southern African &Africa &Communists &Congress &West Germany\\ 
Norway &Finland &Portuguese &African &Belgian &Rhodesia &Swedes &OPEC &Japan \\
Finland &Iceland &Luxembourg &Nigeria &Uganda &South African &Portuguese &French &Bonn \\
Eastern European &USA &Africa &South Africa &African &African &Democratic &State &Southwest Asia \\
Denmark &Sweden &Europeans &Guinea &Salazar &South Africa &Spain &c. &Lincoln \\
non-Soviet &Spain &Sahara &Africa &Portuguese &Africans &Socialist &Spain &New World \\
Netherlands &Norway &non-European &Organization &Liberia &Belgian &Italy &London &Mussolini \\
Danish &USSR &Scandinavian &United Nations &US Del &AF &Rome &British &Marxist \\
\bottomrule
\end{tabular}
}
\label{table:portugal}
\end{table}

\subsection{Importance Scores}\label{Role and Person Importances}

The number of times a person is mentioned in documents, and the relations spanned this way with others, can be leveraged as a measure of that person's relevance. To compute an importance score, we reduce the KG to a unipartite graph where an edge is created for each person's pair mentioned within a document, as represented by his/her ``Role''. Then, we run PageRank centrality algorithm \cite{pagerank}, also implemented in Neo4J, to obtain the importance score at a ``Role'' level. In Table \ref{table:role importance top 10}, we present the most prominent 10 ``Roles''. We see that the main US officials are on top because they are senders, receivers, and decision-makers in US foreign policy. We believe that the US president comes after the US Secretary of State because the latter is the focal point of document traffic. For non-US roles, we see those of other important world countries. All these are obvious results, but help validate the methodology. More complex analysis can be similarly done, for example by computing a dynamic score for specific time ranges, which could capture how relevance changes according to specific events. A similar analysis of ``Person'' importance can be found in Appendix D.


\begin{table}[htb]
\caption{Top 10 most important roles, and corresponding countries in KG-FRUS.}
\centering
\resizebox{0.8\textwidth}{!}{%
\begin{tabular}{ p{9cm} p{1.3cm} p{1.7cm} }
\toprule
Role & Country & PageRank Importance\\
\midrule
United States Secretary of State & US & 19.33\\
President of the United States & US & 16.07\\
National Security Advisor & US & 11.38\\
General Secretary of the Communist Party of the Soviet Union & Russia & 8.22\\
United States Secretary of Defense & US & 8.14\\
Under Secretary of State for Political Affairs & US & 7.76\\
Prime Minister of the United Kingdom & UK & 7.66\\
Secretary of State for Foreign and Commonwealth Affairs & UK & 7.36\\
Chairman of the Joint Chiefs of Staff & US & 6.88\\
President of the French Republic & France & 6.84\\
\bottomrule
\end{tabular}
}
\label{table:role importance top 10}
\end{table}


\section{Conclusions and Future Work}\label{conclusions} 

KG-FRUS is a novel graph-based dataset that covers over one hundred years of US diplomatic relations. Besides the described construction process and paradigm applicable to other corpora in varied fields, we have presented different usages and methods to tackle diverse research questions. Many of these research questions could not be directly answered by the data in text form. The addressed questions are only the tip of the iceberg for a flexible knowledge graph. The proposed questions allow us to showcase the patterns and insights that could be unveiled from the data in scale, thanks to the structure encoded in the KG. {Yet, we need to stress that KG-FRUS may contain errors, which may lead to wrong or biased conclusions, especially when studying really specific events and/or international actors, instead of more global phenomena. Therefore, researchers harnessing KG-FRUS need to be acquainted with this and backtrack the results to the original sources before making concrete solid statements, specially when personal names are involved.}

The validation results presented underpin the methods devised for the steps of person unification and \textit{wikification}. The lack of annotation in the FRUS volumes complicates the expanding number of people nodes in this data which remains an open research question. In future work, we will streamline Named entity recognition to extract most people mentioned in the documents, and relate them to already annotated ones. 


\bibliographystyle{ieeetran}
\bibliography{neurips_2023.bib}


\end{document}